\documentclass[letterpaper]{article} 
\PassOptionsToPackage{table}{xcolor} 
\usepackage{aaai2026}
\usepackage{times}  
\usepackage{helvet}  
\usepackage{courier}  
\usepackage[hyphens]{url}  
\usepackage{graphicx} 
\usepackage{natbib}  
\usepackage{caption} 
\usepackage{algorithm}
\usepackage{algorithmic}
\usepackage[T1]{fontenc}
\usepackage{xcolor}
\usepackage{booktabs}
\usepackage{multirow}
\usepackage{multicol}

\newcommand{\cellc}{\cellcolor{violet!10}}
\usepackage{newfloat}
\usepackage{listings}
\DeclareCaptionStyle{ruled}{labelfont=normalfont,labelsep=colon,strut=off} 
\floatstyle{ruled}
\newfloat{listing}{tb}{lst}{}
\floatname{listing}{Listing}
\usepackage{amsmath}
\usepackage{amsfonts}

\DeclareMathOperator*{\argmax}{arg\,max}
\usepackage{pifont}
\usepackage{dsfont}

\title{STELAR-VISION: Self-Topology-Aware Efficient Learning for \\ Aligned Reasoning in Vision}
\author{
    Chen Li,
    Han Zhang,
    Zhantao Yang,
    Fangyi Chen,
    Zihan Wang,
    Anudeepsekhar Bolimera,
    Marios Savvides
}

\affiliations{
    Carnegie Mellon University\\
    \{chenli4, hanz3, zhantaoy, fangyic, zihanw4, abolimer, marioss\}@andrew.cmu.edu
}

\usepackage{bibentry}

\usepackage{etoolbox} 

\makeatletter
\patchcmd{\@footnotetext}{\parindent}{0pt}{}{}
\makeatother

\begin{document}
\maketitle
\renewcommand{\thefootnote}{}




\footnotetext{%
\hspace*{-1.8em}%
This paper has been accepted at AAAI 2026. This is the author's extended version. The final version will appear in the official proceedings. \\
Project and Datasets:
\textcolor{magenta}{\url{https://stellar-neuron.github.io/stelar-vision/}}%
}


\begin{abstract}
Vision-language models (VLMs) have made significant strides in reasoning, yet they often struggle with complex multimodal tasks and tend to generate overly verbose outputs. A key limitation is their reliance on chain-of-thought (CoT) reasoning, despite many tasks benefiting from alternative topologies like trees or graphs. To address this, we introduce STELAR-Vision, a training framework for topology-aware reasoning. At its core is TopoAug, a synthetic data pipeline that enriches training with diverse topological structures. Using supervised fine-tuning and reinforcement learning, we post-train Qwen2VL models with both accuracy and efficiency in mind. Additionally, we propose Frugal Learning, which reduces output length with minimal accuracy loss.
On MATH-V and VLM\_S2H, STELAR-Vision improves accuracy by 9.7\% over its base model and surpasses the larger Qwen2VL-72B-Instruct by 7.3\%. On five out-of-distribution benchmarks, it outperforms Phi-4-Multimodal-Instruct by up to 28.4\% and LLaMA-3.2-11B-Vision-Instruct by up to 13.2\%, demonstrating strong generalization. Compared to Chain-Only training, our approach achieves 4.3\% higher overall accuracy on in-distribution datasets and consistently outperforms across all OOD benchmarks. We've released datasets, and code will be available.
\end{abstract}


\section{Introduction}
\label{introduction}
Recent advances in large language models (LLMs) have significantly improved reasoning capabilities, with models like GPT-o3 achieving strong performance on complex mathematical and scientific tasks. This progress has extended into the multimodal domain through vision-language models (VLMs) such as GPT-4o~\citep{openai2024gpt4ocard}, GPT-4o-mini~\citep{openai2024o4mini}, and Qwen2.5-VL~\citep{bai2025qwen25vltechnicalreport}. Despite the recent advances, there is still room of improvement in open-sourced VLMs when tackling complex vision-based reasoning tasks (e.g., math and science questions), and the path to enhance their abilities under an affordable training budget remains under-explored.



\begin{figure*}[t]
    \centering
    \includegraphics[width=1\linewidth]{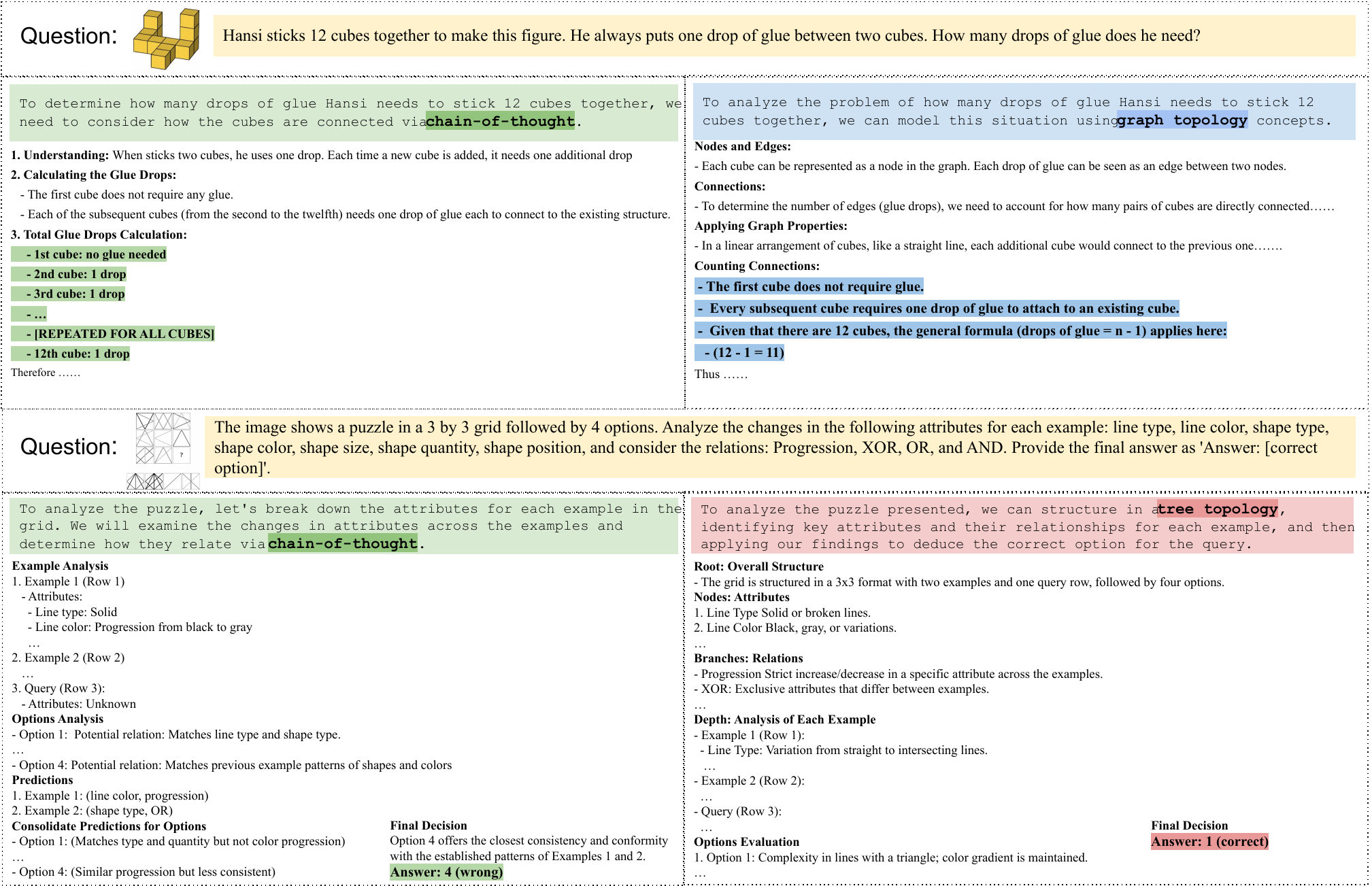}
    \captionsetup{font=small}
    \caption{\textbf{Limitations of the Popular Chain-of-Thought Reasoning Structures.} The widely adopted Chain-of-Thought (CoT) reasoning paradigm (in green) often results in unnecessarily verbose reasoning processes, as demonstrated in the first example. Under CoT reasoning, the model redundantly counts each cube, whereas with $Graph$ topology (in blue), it quickly identifies the key point of the question. In the bottom-row example, CoT reasoning begins with a detailed examination of each subplot but ultimately arrives at an incorrect answer. In contrast, $Tree$ topology (in red) initiates reasoning with a high-level overview before delving into specific features. In both scenarios, CoT-style reasoning proves suboptimal.}
    \label{fig:fig1}
\end{figure*}

To address this, we begin by analyzing VLMs' reasoning behaviors and find that the popular models, both open-source and closed-source, tend to default to the chain-of-thought (CoT)~\citep{wei2023chainofthoughtpromptingelicitsreasoning} generation. This behavior reflects the prevailing trend in their training data, which is overwhelmingly dominated by CoT-style reasoning samples. However, our empirical analysis reveals that \textit{different questions benefit from different reasoning topologies, such as Chain, Tree, or Graph structures }(Please see Figure~\ref{fig:fig1}). The benefits of diverse reasoning topologies have yet been well studied or effectively incorporated into existing training pipelines. Moreover, even state-of-the-art reasoning VLMs tend to generate unnecessarily verbose responses, i.e., ``overthinking", which increases the computational cost and makes real-time applications less viable. We find that there is a correlation between the topological reasoning structures and the output sequence length, thus providing an insight to the overthinking problem created by the CoT reasoning.

We introduce \underline{S}elf-\underline{T}opology-Aware-\underline{E}fficient-\underline{L}earning for \underline{A}ligned
\underline{R}easoning in Vision, \textbf{STELAR-Vision}, a training framework for topology-aware vision-language reasoning. The central to this framework is \textbf{TopoAug}, a synthetic data generation pipeline that produces augmented reasoning topologies with diverse structures, including $\{Chain, Tree, Graph\}$. Specifically, we generate a set of question-answer responses, where each question is repeatedly answered by different topological reasoning structures. For each question, we assign a preferred topology through an automated annotation process.
The generated responses and their labels are used to post-train VLMs via supervised fine-tuning (SFT) and Reinforcement Learning (RL) \citep{meng2024simposimplepreferenceoptimization}.
Regarding the learning process, we observe that RL further amplifies the performance gains introduced by topological augmentation. Moreover, we propose to further increase efficiency with Frugal Learning by promoting shorter responses. We show that only by leveraging the benefits of augmented reasoning topologies, can the model generate concise responses and maintain high accuracy with minimal performance drops, while the counterpart model trained only on CoT-style data fails to gain efficiency without incurring a greater loss in accuracy. 

The augmented topologies expand the exploration space, allowing RL to discover higher-quality optima and leading to superior performance improvement. 
In contrast, the model trained with CoT-style data only sees diminishing returns
We also test our models on five out-of-distribution datasets and consistently outperform its base model. which suggests strong generalization of our proposed STELAR-Vision.  

\noindent Our contributions are summarized as follows:
\begin{itemize}

\item We propose STELAR-Vision, a training framework explicitly trained for topology-aware reasoning. It leverages diverse reasoning topologies such as chains, trees, and graphs, aligns reasoning paths with question characteristics, and enables adaptive and efficient multimodal inference.

\item We introduce TopoAug, a data generation pipeline that automatically produces diverse topological reasoning and annotates optimal structures per question. We also integrate  Frugal Learning into the learning framework, achieving reductions in output length with minimal accuracy tradeoff. 

\item By conducting experiments with post-training supervision and reinforcement learning, STELAR-Vision improves accuracy by 9.7\% over its base model and its larger variant Qwen2VL-72B-Instruct by 7.3\%. On the out-of-distribution dataset, it surpasses 
The Frugal Learning variant reduces output length by 18.1\% while maintaining comparable accuracy.

\end{itemize}

\section{Related Work}
\label{related_work}

\subsection{Topological Reasoning in Language and Vision Models}

Chain-of-Thought (CoT) prompting~\citep{wei2023chainofthoughtpromptingelicitsreasoning} is a widely used reasoning strategy in LLMs and VLMs, guiding models to generate step-by-step solutions. However, its linear structure may not suit all tasks. To address this, Tree-of-Thought (ToT)~\citep{yao2023treethoughtsdeliberateproblem} enables branching exploration, while Graph-of-Thought (GoT)~\citep{Besta_2024} supports iterative and global reasoning. Both improve performance on complex tasks like TSP, algorithmic problem-solving, and multi-stage decision-making.

These methods, however, often rely on rule-based topology generations through sampling and are limited to language-only settings. In contrast, our framework automatically generates diverse topological structures and trains a VLM to adaptively select the optimal one per instance during decoding, enabling more flexible and generalizable reasoning.

\subsection{Reinforcement Learning for LLM and VLM Reasoning}

Reinforcement learning (RL) is a key technique for aligning LLMs and VLMs with desired behaviors in reasoning and preference modeling. Approaches like RLHF~\citep{stiennon2022learningsummarizehumanfeedback, ouyang2022traininglanguagemodelsfollow} and Constitutional AI~\citep{bai2022constitutionalaiharmlessnessai} enable models to acquire complex reasoning and ethical behaviors via preference optimization.

Reward-based methods such as PPO~\citep{schulman2017proximalpolicyoptimizationalgorithms}, RPO~\citep{yin2024relativepreferenceoptimizationenhancing}, and GRPO~\citep{shao2024deepseekmathpushinglimitsmathematical} rely on explicit rewards, while reward-free approaches like DPO~\citep{rafailov2024directpreferenceoptimizationlanguage}, SimPO~\citep{meng2024simposimplepreferenceoptimization}, and ORPO~\citep{hong2024orpomonolithicpreferenceoptimization} achieve comparable results without reward modeling. These methods are widely applied to mathematical reasoning, long-horizon tasks, and instruction tuning.

In VLMs, RL has been used to enhance structured reasoning and safety-critical applications. VLM-RL~\citep{huang2024vlmrlunifiedvisionlanguage} improves decision-making in autonomous driving, MedVLM-R1~\citep{pan2025medvlmr1incentivizingmedicalreasoning} ensures safety in medical imaging, and RLVR~\citep{chen2025rlvrinvlms} boosts OOD generalization in tasks like visual counting and open-ended QA.

Building on these insights, we show that combining topology-aware data generation with RL (e.g., SimPO) improves both accuracy and efficiency. Topological diversity expands the exploration space, increasing the likelihood of discovering stronger reasoning strategies during RL.

\subsection{Curriculum Learning and Structured Reasoning}


Curriculum learning~\citep{bengio2009curriculum} trains models on progressively harder tasks, aiding structured reasoning. Recent works extend this idea: Xi et al.~\citep{xi2024traininglargelanguagemodels} use reverse curricula for LLMs, Zhao et al.~\citep{zhao2024automaticcurriculumexpertiteration} propose Auto-CEI for ability-aligned sampling, and Ma et al.~\citep{ma2025problemsolvinglogicguidedcurriculum} design logic-guided curricula for in-context learning.

In VLMs, LlamaV
o1~\citep{thawakar2025llamavo1rethinkingstepbystepvisual} schedules visual examples for step-wise reasoning, and VLM-R1~\citep{shen2025vlmr1} integrates curriculum-aware GRPO for stability and interpretability. However, most curricula are static and predefined.

In contrast, our approach is an end-to-end learning via a dataset from TopoAug, where synthetic reasoning paths evolve with model performance. Coupled with RL and topological diversity, our training framework promotes structured reasoning in a more scalable and generalizable way.

\section{Method}
\label{sec:method}
\begin{figure}[t]
    \centering
    \includegraphics[width=1\linewidth]{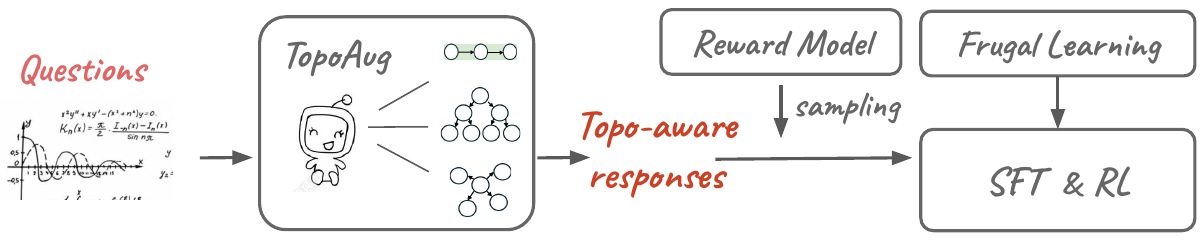}
    \caption{An overview of the STELAR-Vision framework }
    \label{fig:methodpipe}
\end{figure}

In this section, we first construct topology-aware responses on two mathematical datasets. We then investigate the relationship between the topological reasoning and response accuracy. Finally, we present the topology-aware training framework shown in Figure \ref{fig:methodpipe}.

\subsection{Constructing Topology-Aware Responses}


\subsubsection{Data}
We begin with two mathematical datasets MathVision (MATH-V) \cite{wang2024measuring}, and VLM\_S2H \cite{park2025generalizing}. They provide diverse samples requiring high-level reasoning, suitable to our study. Every sample in each dataset is composed of an image and a text-based problem description. MATH-V contains 3,040 math problems with visual contexts from real-world competitions. VLM\_S2H~\cite{park2025generalizing} offers 7,000 structured reasoning puzzles testing logical relations like Progression and Exclusive OR.

\subsubsection{TopoAug: Generating Topology-Aware Response} 
For each question, we prompt these models to repeatedly generate multiple responses across three reasoning topologies, $T = \{Chain, Tree, Graph\}$ with extensive degrees of freedom in maximum depth, number of children, and number of neighbors. Please see the supplementary for detailed prompts.

To ensure diversity in the reasoning topologies and to maintain a balanced distribution of positive and negative samples in our dataset, we use two models with different scale and abilities to fulfill the generation: open-sourced Qwen2-VL-7B-Instruct \citep{wang2024qwen2vlenhancingvisionlanguagemodels}, and commercial GPT-4o-Mini~\citep{openai2024gpt4o} for data generation. 

Each question gets multiple generated responses via different topology prompting.

\paragraph{Topology Label and Outcome Label} 
We assign each question three Topology Label $\mathcal{F}_{q, t}$ and assign each response an Outcome Label $\mathcal{H}_r$, which are defined as below.

\begin{itemize}
    \item Topology Label $\mathcal{F}_{q, t}$: A continuous value in the range \([0,1]\), representing the accuracy that a given topology produces the correct answers for a question. This question-specific label reflects how effectively each reasoning topology solves a given problem. For each problem \(q\), we compute the accuracy of the responses from each topology type and assign it as: \(\mathcal{F}_{q, t} =  \frac{N_{\text{correct}}(q, t)}{N_{\text{total}}(q, t)}\), where \(N_{\text{correct}}(q, t)\) is the number of correct responses using topology \(t\) for question \(q\), and \(N_{\text{total}}(q, t)\) is the total number of responses generated using \(t\).
    \item Outcome Label \(\mathcal{H}_r\): For each generated response, a binary value \(\{0,1\}\) indicating whether it is correct. Each response \(r\) is assigned label 1 if correct and 0 otherwise.
\end{itemize}

\paragraph{Problems Difficulty Segmentation} 
TopoAug enables us to quantify problem to three difficulty levels based on the distribution of topology labels:

\begin{itemize}
    \item Easy: Problems where all three topology scores exceed a quantile threshold (85\%).
    \item Hard: Problems where all three topology scores fall below a specified quantile threshold (15\%) in their respective distributions.
    \item Medium: Problems that do not meet the criteria for either hard or easy categories.
\end{itemize}

\subsection{Analysis: Topological Reasoning Structures}
We first show that different questions are better solved using distinct reasoning topologies. 
We evaluate the two aforementioned vision-language models (VLMs). We exclude the newer \textsc{Qwen2.5-VL}\cite{qwen2025qwen25technicalreport} series from our analysis due to its relatively unstable performance in generating diverse reasoning topologies.

We first prompt the models to generate responses using their default reasoning behavior. Next, we explicitly instruct them to reason using three distinct topologies: \textit{Chain}, \textit{Tree}, and \textit{Graph}. We compute a topology-wise \textit{Win Rate} to assess the performance of each topological reasoning structure, which is defined below. We then conduct a subject category-wise study on win rate. 

\textbf{Win Rate}: We measure across the entire dataset and calculate the percentage of occurrence where a topology $t$ is the best performing reasoning structure among the three topology types in Equation~\ref{eq:winrate}:

\begin{equation}
    \text{Win Rate}(t) = \frac{\sum_{q \in Q} \mathds{1}_t ( \argmax_{t' \in T} \mathcal{F}_{q, t'})}{N_{Q}}
    \label{eq:winrate}
\end{equation}

where \( N_Q \) is the total number of questions, and  $\mathds{1}$ is the indicator function. For each question, the topology with the highest Topology Label $\mathcal{F}_{q, t}$ is assigned a win. We first analyze which topological reasoning structure works the best for the questions. The Win Rate statistics is presented in Table~\ref{tab:2}.

\begin{table}[htbp]
    \centering
    \resizebox{0.6\columnwidth}{!}{%
    \begin{tabular}{lccc}
        \toprule
        \textbf{} & Chain & Tree & Graph \\
        \midrule
        Win Rate & 49\% & 28\% & 23\% \\
        \bottomrule
    \end{tabular}
    }
    \captionsetup{font=small}
    \caption{Comparison of Win Rates across reasoning topologies.}
    \label{tab:2}
\end{table}

\begin{figure}[ht]
    \centering
    \includegraphics[width=1.0\linewidth]{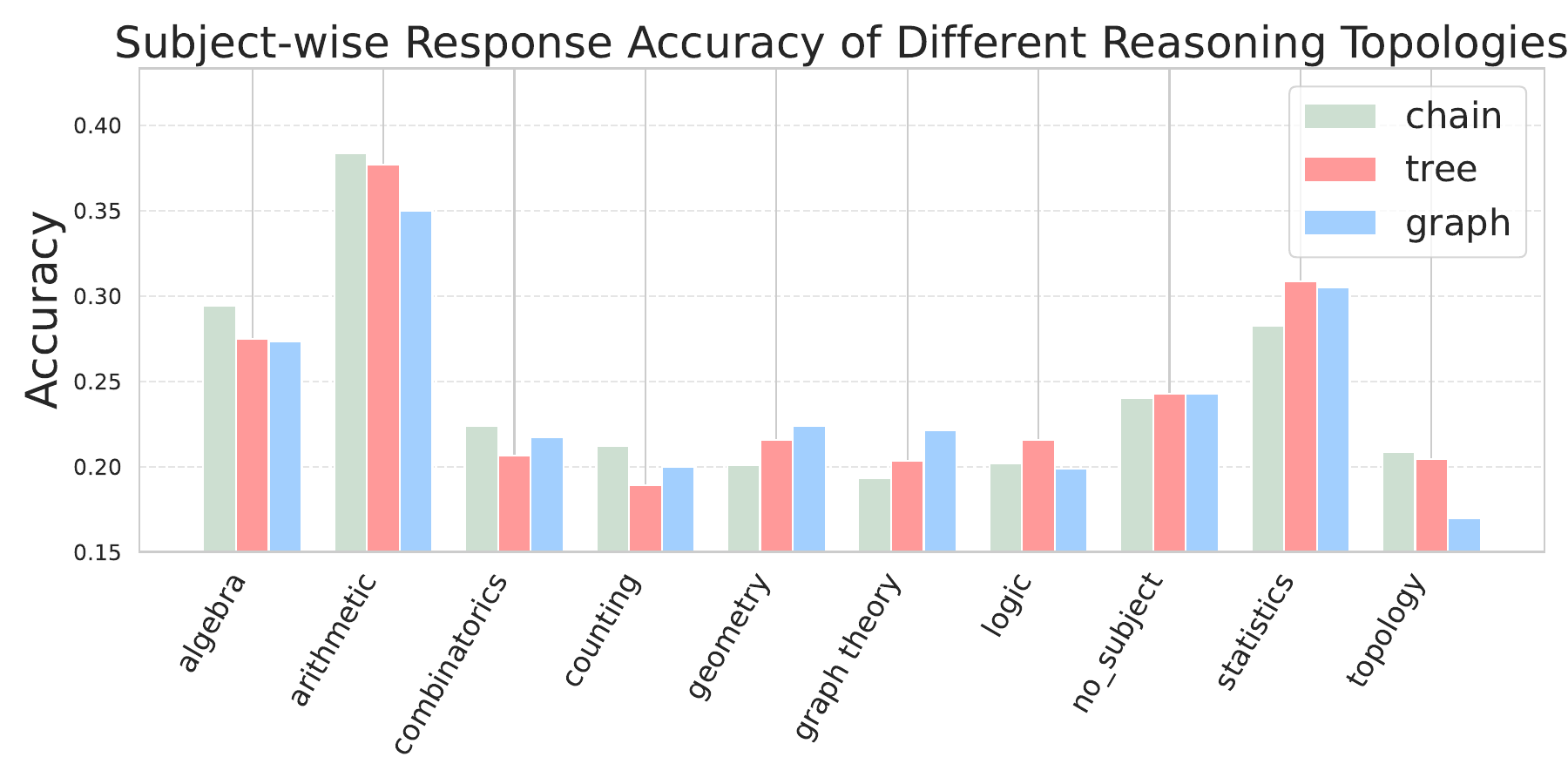}  
    \captionsetup{font=small}
    \caption{\textbf{Comparison of topology accuracy across subjects:} Accuracy of $Chain$, $Tree$, and $Graph$ reasoning topological structures per subject of MATH-V dataset. $Chain$ remains the best overall reasoning structure, while $Tree$, and $Graph$ perform better in at reasoning subjects such as ``graph theory" or ``statistics".}
    \label{fig:topo_subjects}
\end{figure}

Our analysis in Table \ref{tab:2} reveals that $Chain$ structure reasoning remains the best overall reasoning pattern in both VLMs, regardless of model size. However, $Tree$ and $Graph$ collectively account for more than half of the winning responses. We conduct a detailed investigation of each topology's performance across subject domains by following the 10 subject categories in the MATH-V dataset and measuring the accuracy of each reasoning topology per subject, as illustrated in Figure \ref{fig:topo_subjects}. 
And thus we arrive at a conclusion that different problems benefit from different topologies, and identifying the optimal ones yields accuracy gains. This aligns with theoretical expectations—while straightforward tasks are adequately addressed by CoT-style reasoning, more complex or structurally intricate problems require richer topological representations to capture their underlying relationships effectively.

\begin{figure}[h]
    \centering
    \includegraphics[width=1.0\linewidth]{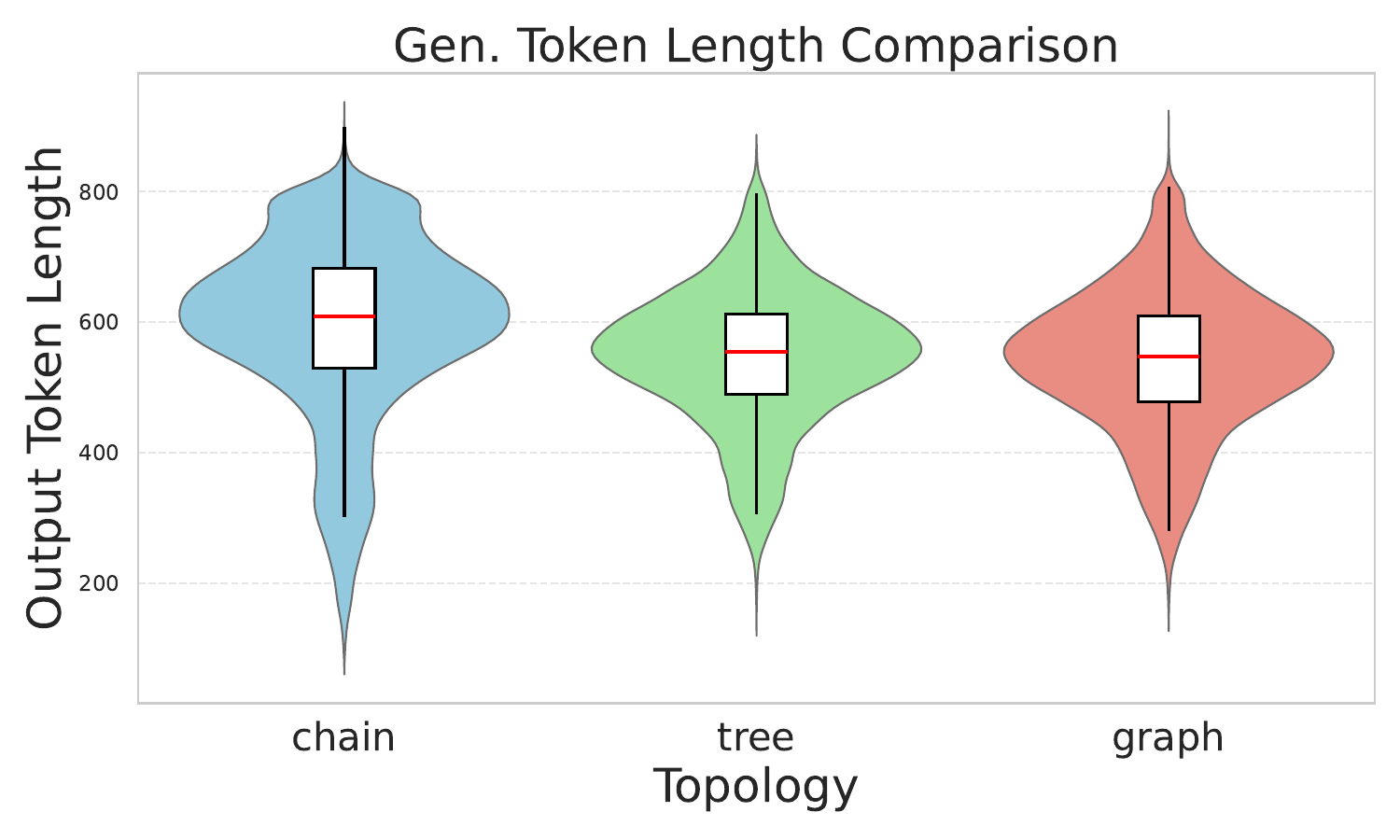}  
    \captionsetup{font=small}
    \caption{Distribution of generated reasoning token length of $Chain$, $Tree$, and $Graph$ topological structures in TopoAug Dataset. The box within each violin plot represents the median, and 25\% and 75\% percentile thresholds.}
    \label{fig:gen_len}
\end{figure}

\textbf{Topology-Wise Generation Length} We further investigate the reasoning generation token length distribution within the TopoAug dataset. As illustrated in Figure \ref{fig:gen_len}, while most generations exhibit an average length of approximately 550 tokens, the $Chain$ topology produces the longest generations with a right-skewed distribution that favors extended reasoning processes. In contrast, both $Tree$ and $Graph$ topologies demonstrate similar token length distributions with overall a lower length, indicating comparable reasoning verbosity across these structured approaches.

\subsection{STELAR-Vision Post-Training}
Our finding propels us to make two reasonable assumptions:
\begin{itemize}
\item \textbf{Assumption 1:} A model trained with topologically diverse reasoning structures—without increasing data volume—can achieve higher reasoning accuracy by learning to adaptively identify the best reasoning structure for each problem at test time.
\item \textbf{Assumption 2:} Building up Assumption 1, a learning mechanism can be designed to encourage concise yet accurate outputs, enhancing inference efficiency with minimal performance loss.
\end{itemize}
We device a Post-training framework that consists of two phases: Supervised Fine-Tuning (SFT) and Reinforcement Learning (RL).



\subsubsection{Phases 1: Supervised Fine-tuning}
\label{sec:sft}
We collect the data generated via TopoAug and combine it with three additional datasets, OKVQA\cite{marino2019okvqavisualquestionanswering}, A-OKVQA\cite{schwenk2022aokvqabenchmarkvisualquestion}, and LLaVA150k-Instruct \cite{liu2023visualinstructiontuning}. We include them as general VQA tasks to preserve the model’s basic VQA capabilities. Note that the additional 3 VQA datasets are used without any topological augmentation to avoid altering their original structure and maintain generalization. We perform SFT on TopoAug-generated data mixed with general VQA datasets, using a three-step filtering process: (1) balanced sampling from Easy, Medium, and Hard problems, (2) keeping only responses with positive outcome labels, and (3) rejection sampling with a 7B Outcome Reward Model (ORM)~\citep{ouyang2022traininglanguagemodelsfollow}, trained with both topology and outcome labels to select higher-quality samples with greater win potential.

The fine-tuning uses LoRA~\citep{hu2021loralowrankadaptationlarge} with next-token prediction (NTP), minimizing the loss 
\begin{equation}
    \mathcal{L}_{\text{NTP}} = - \sum_{t=1}^{T} \log P_\theta(y_t \mid y_{<t}, x)
\end{equation}
 where $x$ is the input, $y = (y_1, \dots, y_T)$ the target, and $P_\theta$ the model's predicted token probabilities.

\subsubsection{Phases 2: Reinforcement Learning}
We follow prior work~\citep{chu2025sftmemorizesrlgeneralizes, zhai2024finetuninglargevisionlanguagemodels} by initializing reinforcement learning (RL) from an SFT checkpoint, which improves both in-distribution accuracy and OOD generalization. We adopt SimPO~\citep{meng2024simposimplepreferenceoptimization} for its simplicity and alignment with test-time behavior. Its objective encourages preferred responses over less-preferred ones, and is defined as 
\begin{equation}
\begin{split}
\mathcal{L}_{\text{SimPO}}(\pi_\theta) = 
& -\mathbb{E}_{(x, y_w, y_l) \sim \mathcal{D}} \Bigg[ 
    \log \sigma \Bigg( 
        \frac{\beta}{|y_w|} \log \pi_\theta(y_w \mid x) \\
& \qquad\qquad\qquad\quad - \frac{\beta}{|y_l|} \log \pi_\theta(y_l \mid x) 
        - \gamma 
    \Bigg) 
\Bigg]
\end{split}
\end{equation}

where $x$ is the input, $y_w$ and $y_l$ are the preferred and less-preferred responses, $\pi_\theta$ is the policy, $\sigma$ is the sigmoid, and $\beta$, $\gamma$ are temperature and margin parameters.

We compare two RL setups: one trained on TopoAug-based preference pairs and one on Chain-only data of equal size. Correct responses are treated as the preferred responses. To prevent data leakage, we remove topology prompts during training, so the model must infer optimal structures at test time.

\begin{table*}[t]
    \centering
    \resizebox{\textwidth}{!}{%
    \begin{tabular}{@{}l|ccc|ccccc@{}} 
        \toprule[1pt]
        \multirow{2}{*}{Model} &
        \multicolumn{3}{c|}{\textbf{In-Distribution Accuracy (\%)}} &
        \multicolumn{5}{c}{\textbf{Out-of-Distribution Accuracy (\%)}} \\ \cmidrule(l){2-4} \cmidrule(l){5-9} 
        \multicolumn{1}{l|}{} &
        \multicolumn{1}{c}{VLM\_S2H} & 
        \multicolumn{1}{c}{MATH-V} &
        \multicolumn{1}{c|}{Overall} &
        \multicolumn{1}{c}{Geometry3K} &
        \multicolumn{1}{c}{We-Math} &
        \multicolumn{1}{c}{PolyMath} &
        \multicolumn{1}{c}{SciBench} &
        \multicolumn{1}{c}{LogicVista} \\
        \midrule
        GPT-4o \cite{openai2024gpt4o}  & 32.0 & 28.0 & 30.7 & 57.0 & 66.4 & 25.0 &  31.1 & 34.6 \\
        LLaVA-v1.6-Mistral-7B \cite{liu2024llavanext}  & 26.0 & 8.0 & 18.0 & 20.6 & 26.0 & 9.2 &  3.4 & 18.5 \\
        Llama-3.2-11B-Vision-Instruct \cite{grattafiori2024llama3herdmodels,meta2024llama32vision} & 22.0 & 10.0 & 18.0 & 35.0 & 37.8 & 22.2 & 10.7 & 24.8 \\
        MiniCPMv2.6-8B \cite{yao2024minicpm}  & 1.5 & 13.0 & 18.7 & 45.0 & 50.2 & 14.4 & 8.5 & 20.7 \\
        Phi-4-multimodal-5.6B-instruct \cite{abouelenin2025phi}  & 23.0 & 11.0 & 22.0 & 8.4 & 35.8 & 10.2 & 10.2 & 6.7 \\
        InternVL3-9B \cite{zhu2025internvl3}  & 25.0 & 21.0 & 27.3 & 41.2 & 51.4 & 21.6 &  20.3 & 32.6 \\
        Qwen2VL-72B-Instruct \cite{yang2024qwen2technicalreport} & 21.0 & 20.0 & 20.7 & 50.2 & 60.6 & 13.0 & 25.4 & 28.8 \\
        Qwen2VL-7B-Instruct \cite{yang2024qwen2technicalreport} & 21.0 & 13.0 & 18.3 & 35.2 & 46.6 & 16.0 & 10.7 & 17.0 \\
        \midrule
        \cellc Chain-Only & \cellc 25.0 & \cellc 21.0 & \cellc 23.7 & \cellc 31.4 & \cellc 42.2 & \cellc 17.2 & \cellc 10.7 & \cellc 25.4 \\
        \cellc STELAR-Vision-SFT &\cellc 28.0 &\cellc \textbf{24.0} &\cellc 26.7 &\cellc \textbf{44.4} & \cellc 47.4 &\cellc 24.8 &\cellc 9.0 &\cellc \textbf{33.3} \\
        \cellc STELAR-Vision-RL-ONLY & \cellc 24.0 & \cellc  23.0 & \cellc 23.7 & \cellc 32.8 & \cellc 39.0 & \cellc \textbf{26.0} & \cellc \textbf{17.5} & \cellc 23.9 \\
        \cellc STELAR-Vision &\cellc \textbf{31.0} &\cellc 22.0 &\cellc \textbf{28.0} &\cellc 36.8 &\cellc \textbf{51.0} &\cellc 23.8 &\cellc 12.4 &\cellc 29.0 \\
        \bottomrule
    \end{tabular}
    }
    \captionsetup{font=small}
    \caption{\textbf{Quantitative Evaluation.} STELAR-Vision achieves strong gains across both in-distribution and out-of-distribution reasoning benchmarks. On ID datasets, it outperforms its base model Qwen2VL-7B-Instruct by \textbf{9.7\%}, and even surpasses the larger Qwen2VL-72B-Instruct by \textbf{7.3\%}. On OOD benchmarks, it exceeds Phi-4-multimodal-instruct by up to \textbf{36\%} and LLaMA-3.2-11B-Vision-Instruct by up to \textbf{13.2\%}. Compared to Chain-Only training, STELAR-Vision achieves up to \textbf{13\%} higher accuracy, highlighting the power of topological augmentation.}
    
    \label{tab:sota}
\end{table*}

\paragraph{Frugal Learning}
\label{sec:fugal}
While recent work has explored efficient reasoning in LLMs~\citep{arora2025traininglanguagemodelsreason, aggarwal2025l1controllinglongreasoning}, these approaches rely on additional reward models, lack cost-controllability, and do not focus on vision-based reasoning. To improve test-time reasoning efficiency, we propose \textit{Frugal Learning}, which trains a compact variant called STELAR-V-Short. We propose and compare two training strategies: 

\textbf{Variant 1: STELAR-Vision-Short$^{\dagger}$.} We first introduce a filter that identifies ``short and correct" responses as preferred targets, where the outcome Label $\mathcal{H}_r = 1$ and the generated token length falls below a 25\% percentile threshold. The model undergoes training with both supervised fine-tuning (SFT) and reinforcement learning (RL). During RL training, we designate ``short and correct" responses as winners and incorrect responses as losers in the preference optimization process. We also train a model with Chain-Only data with the same process and we denote it as \textbf{Chain-Only-Short$^{\dagger}$}.

\textbf{Variant 2: STELAR-Vision-Short$^{\ddagger}$.} A natural alternative approach to encourage shorter and more efficient responses involves explicitly penalizing lengthy outputs. Building upon STELAR-Vision-Short$^{\dagger}$, we investigate an alternative configuration, STELAR-Vision-Short$^{\ddagger}$, which maintains the preference for ``short and correct" responses while treating both ``incorrect responses" and ``correct yet lengthy" responses as losers during RL preference training. This design encourages the model to avoid excessive token generation while maintaining accuracy in reaching the correct final answer.



\section{Experiments}
\label{exp}

\subsection{Experimental Setup}
\label{exp_setup}

\paragraph{Datasets}
We use In-Distribution (ID) datasets from the Method Section~\ref{sec:method}, splitting them into train/test sets with ~50K–60K samples for training. OKVQA and A-OKVQA provide 18k and 36k VQA pairs, with A-OKVQA adding rationales and new knowledge types. We also use 17k instruction-following examples from LLaVA-150k for multimodal training. The number of the training samples in Table~\ref{tab:training_data}. 


\begin{table}[htbp]
    \centering
    \resizebox{1.0\columnwidth}{!}{%
    \begin{tabular}{l|cccccc}
        \toprule
        Dataset & MATH-V & VLM\_S2H & OKVQA & A-OKVQA & LLava150k-inst \\
        \midrule
        Sample & 85K    & 160K    & 18K    & 20K     & 17K \\
        \bottomrule
    \end{tabular}}
    \captionsetup{font=small}
    \caption{Total sample sizes of datasets used for training.}
    \label{tab:training_data}
\end{table}

Evaluation is performed on the test sets of MATH-V and VLM\_S2H. For Out-of-Distribution (OOD) evaluation, we use five recent benchmarks with three math oriented datasets, Geometry3K \cite{DBLP:journals/corr/abs-2105-04165}, We-Math \cite{qiao2024we} and PolyMath \cite{gupta2024polymath}, a STEM dataset SciBench \cite{wang2024scibench} and a generic logic reasoning dataset LogicVista \cite{xiao2024logicvistamultimodalllmlogical}. In Table \ref{tab:eval_dataset}, we show the evaluation datasets as well as their respective sizes. Collectively, we gather 2,425 evaluation samples. 

\paragraph{Models}
Our base model is Qwen2VL-7B-Instruct~\citep{wang2024qwen2vlenhancingvisionlanguagemodels}, chosen for its stable generation of all three topologies in 74\% of cases. We exclude the newer Qwen2.5VL-7B-Instruct~\citep{qwen25vl7b} for its instability in producing $Tree$ and $Graph$ structure reasoning.
We also compare our models with additional open-source and proprietary models. 

\begin{table}[h!]
\centering
\resizebox{\columnwidth}{!}{%
\begin{tabular}{rcccc}
\toprule
Dataset & Subject & Is OOD? & Question Type & Sample Size \\
\midrule
VLM\_S2H & Math & \ding{55} & multiple-choice & 200 \\
MATH-V & Math & \ding{55} & free-form, multiple-choice & 100 \\
Geometry3K & Math & \ding{51} & multiple-choice & 500 \\
We-Math & Math & \ding{51} & multiple-choice & 500 \\
PolyMath & Math & \ding{51} & multiple-choice & 500 \\
SciBench & STEM & \ding{51} & free-form & 177 \\
LogicVista & Generic & \ding{51} & multiple-choice & 448 \\
\midrule
Total & & & &  2425 \\
\bottomrule
\end{tabular}
}
\captionsetup{font=small}
\caption{Summary of Evaluation Datasets}
\label{tab:eval_dataset}
\end{table}

\paragraph{Evaluation Metrics}
We report \textbf{Accuracy} as the main metric: \(\text{Accuracy} = \frac{\sum_{q \in Q} \text{Answer}(q) = \text{GT}(q)}{N_Q}\), where GT stands for the ground-truth answer. To assess efficiency under Frugal Learning, we also consider \textbf{Generated Token Length} as a metric for reasoning efficiency.

All experiments are conducted on eight NVIDIA A100/H100 GPUs with 80GB of memory each. Supervised fine-tuning (SFT) on a 7B model with ~50K–60K samples takes approximately 5–7 hours, while reinforcement learning (RL) on the same scale requires 8–10 hours, depending on system variability.

\subsection{Overall Evaluation Results}
We use STELAR-Vision to denote models trained with both phases of post-training, the -SFT suffix indicates models trained with supervised fine-tuning only, and -RL-ONLY indicates reinforcement learning directly from the base model without SFT.

Table~\ref{tab:sota} shows results on the in-distribution datasets MATH-V and VLM\_S2H as well as OOD datasets. STELAR-Vision achieves the highest in-distribution accuracy, significantly outperforming its base model Qwen2VL-7B-Instruct by \textbf{9.7\%}, and surpassing larger model LLaMA-3.2-11B by \textbf{10.0\%} and Qwen2VL-72B-Instruct by \textbf{7.3\%}.

Results on out-of-distribution (OOD) benchmarks highlight the strong generalization of STELAR-Vision. It outperforms its base model Qwen2VL-7B-Instruct on all five OOD datasets and trails GPT-4o by only 1.2\% on We-Math. Compared to recently popular model Phi-4-Multimodal-instruct, STELAR-Vision achieves significantly higher accuracy across all OOD tasks, including a \textbf{+28.4\%} gain on the spatial reasoning benchmark Geometry3K. It also surpasses other strong open-source models such as LLaVA-v1.6, LLaMA-3.2-11B-Vision, MiniCPMv2.6-8B, and InternVL3-9B on most benchmarks, demonstrating the impact of our TopoAug-enhanced training. 

Interestingly, the \texttt{-SFT} and \texttt{-RL-ONLY} variants sometimes outperform the full model. We attribute this to: (1) \texttt{-SFT} potentially overfitting due to memorization~\cite{chu2025sftmemorizesrlgeneralizes}, and (2) \texttt{-RL-ONLY} occasionally lacking alignment. Topological augmentation appears to mitigate both limitations by improving generalization and alignment.

Finally, we find that Chain-Only models, despite moderate in-distribution gains, fall short on OOD tasks, further validating the necessity of topology-aware training. See Section~\ref{sec:ablation} for details.


\subsection{Ablation Studies}
\label{sec:ablation}

We perform ablation studies comparing our models to counterparts trained solely on chain-based reasoning data. As shown in Table~\ref{tab:ablation1}, \textbf{STELAR-Vision} consistently outperforms Chain-Only variants on in-distribution (ID) datasets, improving accuracy from 23.7\% to 28\% (+\textbf{4.3\%}). Table~\ref{tab:sota} further confirms this trend across benchmarks. On out-of-distribution (OOD) datasets, \textbf{STELAR-Vision} achieves up to \textbf{8.8\%} higher accuracy than Chain-Only models.

These findings collectively demonstrate that \textbf{STELAR-Vision} benefits not only from data distillation or pattern memorization but also from a genuine ability to internalize and adaptively select optimal reasoning topologies based on the task.

\begin{table}[htbp]
\centering
\resizebox{\columnwidth}{!}{%
\begin{tabular}{l|cc|ccc}
\toprule
\textbf{Model} & \textbf{w/ SFT} & \textbf{w/ RL} & \textbf{VLM\_S2H} & \textbf{MATH-V} & \textbf{Overall} \\
\midrule
Qwen2VL-7B-Instruct & & & 21.0 & 13.0& 18.3 \\
\midrule
Chain-Only-SFT & \ding{51} & & 18.5 & 19.0 & 18.7   \\
Chain-Only-RL-ONLY & & \ding{51} & 23.5 & 15.0  & 20.7  \\
Chain-Only & \ding{51} & \ding{51} & 25.0 & 21.0 & 23.7   \\
STELAR-Vision-SFT & \ding{51} & & 28.0 &  24.0 & 26.7  \\
STELAR-Vision-RL-ONLY & & \ding{51} & 24.0 & 23.0 & 23.7  \\
STELAR-Vision & \ding{51} & \ding{51} & \textbf{31.0} & \textbf{22.0} & \textbf{28.0}  \\
\bottomrule
\end{tabular}
}
\captionsetup{font=small} 
\caption{\textbf{Impact of TopoAug Dataset and Training Methods.} We present an ablation study on the in-distribution VLM\_S2H and Math-V datasets to compare the performance of our models against counterparts trained exclusively on chain-based reasoning data across all training methods. \textbf{STELAR-Vision} consistently outperforms all Chain-Only variants across all ID datasets-specifically it improves the highest variant Chain-Only from 25\% to 31\% by \textbf{6\%}, and boosts overall accuracy by \textbf{4.3\%}, highlighting the effectiveness of topological augmentation.}
\label{tab:ablation1}
\end{table}


\subsection{Efficiency Gains from Frugal Learning}

We assess the impact of Frugal Learning on both in-distribution (ID) and out-of-distribution (OOD) performance, as shown in Table~\ref{tab:accuracy_length}. 
The difference between STELAR-Vision-Short$^{\dagger}$ and STELAR-Vision-Short$^{\ddagger}$ is detailed in Section~\ref{sec:fugal}.




As shown in Table~\ref{tab:accuracy_length}, STELAR-Vision-Short$^{\dagger}$ reduces token length by 101 (ID) and 24.5 (OOD), while still outperforming Qwen2VL-7B-Instruct by 2.5\%. However, it shows a 2.9\% accuracy drop compared to the full STELAR-Vision. In contrast, STELAR-Vision-Short$^{\ddagger}$ yields inconsistent length reduction and greater performance loss, likely due to conflicting optimization signals that penalize correct outputs.

Chain-Only-Short$^{\dagger}$ also shortens outputs but suffers from even larger accuracy degradation. Notably, Chain-Only models fine-tuned with RL often generate overly verbose responses, even with Frugal Learning—highlighting the advantage of TopoAug in promoting concise and effective reasoning.

\begin{table}[htbp]
    \centering
    \scriptsize  
    \begin{tabular}{@{}lccc@{}}
        \toprule
        \multirow{2}{*}{\textbf{Model}} &
        \multirow{2}{*}{\textbf{Accuracy (\%)}} & \multicolumn{2}{c}{\textbf{Gen. Token Length}} \\
        \cmidrule(l){3-4}
        & & ID & OOD  \\
        \midrule
        Qwen2VL-7B-Instruct & 26.2 & 613.5 & 543.3 \\
        \midrule
        Chain-Only & 28.7 & 878.4 & 742.6 \\
        Chain-Only-Short$^{\dagger}$ & 23.9 & 843.1 & 713.0 \\
        STELAR-Vision-SFT & 26.7 & 604.4 & 483.3\\
        STELAR-Vision & 31.6 & 556.7 & 523.4 \\
        \midrule
        STELAR-Vision-Short$^{\dagger}$ & 28.7 & \textbf{455.7} & \textbf{498.6} \\
        STELAR-Vision-Short$^{\ddagger}$ & 21.9 & 538.9 & 555.9\\
        \bottomrule
    \end{tabular}
    \vspace{2pt}
    \captionsetup{font=small}
    \caption{\textbf{Comparison of accuracy and generated token length across models:} STELAR-Vision improves performance while using fewer generation tokens. Frugal learning further improves generation efficiency.}
    \label{tab:accuracy_length}
\end{table}


\begin{table}[htbp]
    \centering
    \scriptsize  
    \begin{tabular}{c|c|ccc}
        \toprule
        Model & Dataset & Tree & Graph & Chain \\
        \midrule
        w/o STELAR-Vision & Overall & - & - & 100.00 \\
        \midrule
         \multirow{6}{*}{w/ STELAR-Vision} & ID & 14.3 & 9.7 & 76.0 \\
         & We-Math & 63.0 & 7.4 & 29.6 \\
         & Geometry3K & 96.4 & 3.0 & 0.6 \\
         & LogicVista & 22.7 & 15.6 & 61.7 \\
         & PolyMATH & 54.0 & 14.8 & 31.2 \\
         & SciBench & 54.2 & 23.2 & 22.6 \\
        \bottomrule
    \end{tabular}
    \captionsetup{font=small}
    \caption{\textbf{Impact of Training on Test-time Topology Selection.} Percentage of reasoning topologies autonomously selected by each model on our evaluation datasets, without explicit prompting. ID denotes the in-distribution test split.}
    \label{tab:topology_fraction}
\end{table}

\subsection{Why Our Method Works?}

Our method achieves substantial improvements on both in-distribution and out-of-distribution benchmarks. We hypothesize two key reasons for this success. First, for models that already possess the ability to generate diverse reasoning topologies but lack the capability to select the optimal topology for a given problem (e.g., Qwen2VL-7B-Instruct), our approach enables the model to adaptively choose the most effective structure. Second, for models that might not inherently support diverse topologies, our framework could instill this ability through guided supervision and reinforcement learning, and thus our methodology is still universal. However, due to computational constraints, we have not yet conducted a systematic study to isolate this effect, which we leave for future work.


Empirical results support these hypotheses: STELAR-Vision consistently outperforms the Chain-only baseline across both in-distribution and out-of-distribution tasks, as shown in Table~\ref{tab:sota}. Moreover, Table~\ref{tab:topology_fraction} reveals a notable post-training increase in the generation of tree and graph structures, which correlates with accuracy gains—demonstrating the utility of topological augmentation.

On out-of-distribution benchmarks, we observe that datasets such as LogicVista favor simpler, more generic reasoning and hence exhibit a higher frequency of chain-based reasoning. In contrast, datasets requiring more complex reasoning, such as Geometry3K and SciBench, show an increased prevalence of tree and graph topologies. This distribution indicates that the model does not merely memorize patterns but has genuinely learned to select the most suitable topology based on the nature of the problem.

\section{Conclusion and Discussion}
\label{conclude}

In this work, we propose STELAR-Vision, a training framework that enables topology-aware reasoning in VLMs via generated responses. STELAR-Vision enhances vision-language reasoning by leveraging diverse topological structures, achieving a 9.7\% accuracy improvement over its base model and outperforming its larger variant Qwen2VL-72B-Instruct by 7.3\%. The Frugal Learning variant reduces output length by 18.1\% while maintaining comparable accuracy, surpassing Chain-Only baselines in both efficiency and task effectiveness. STELAR-Vision demonstrates strong generalization across five diverse OOD datasets and achieves 4.3\% higher overall accuracy on in-distribution tasks, consistently outperforming Chain-Only training.

Despite these promising results, STELAR-Vision currently relies on predefined topology types, and the dynamic relationship between problem structure and optimal reasoning topology remains underexplored. Future work will focus on enabling scalable, end-to-end topology induction and extending applicability to more sophisticated multimodal reasoning tasks.


\bibliography{aaai2026}

\clearpage
\appendix
\section*{Supplementary Material}

\vspace{2.0em} 

\begin{center}
    \textbf{Example Prompts for TopoAug:}

    \vspace{1em} 

    \includegraphics[width=0.95\textwidth]{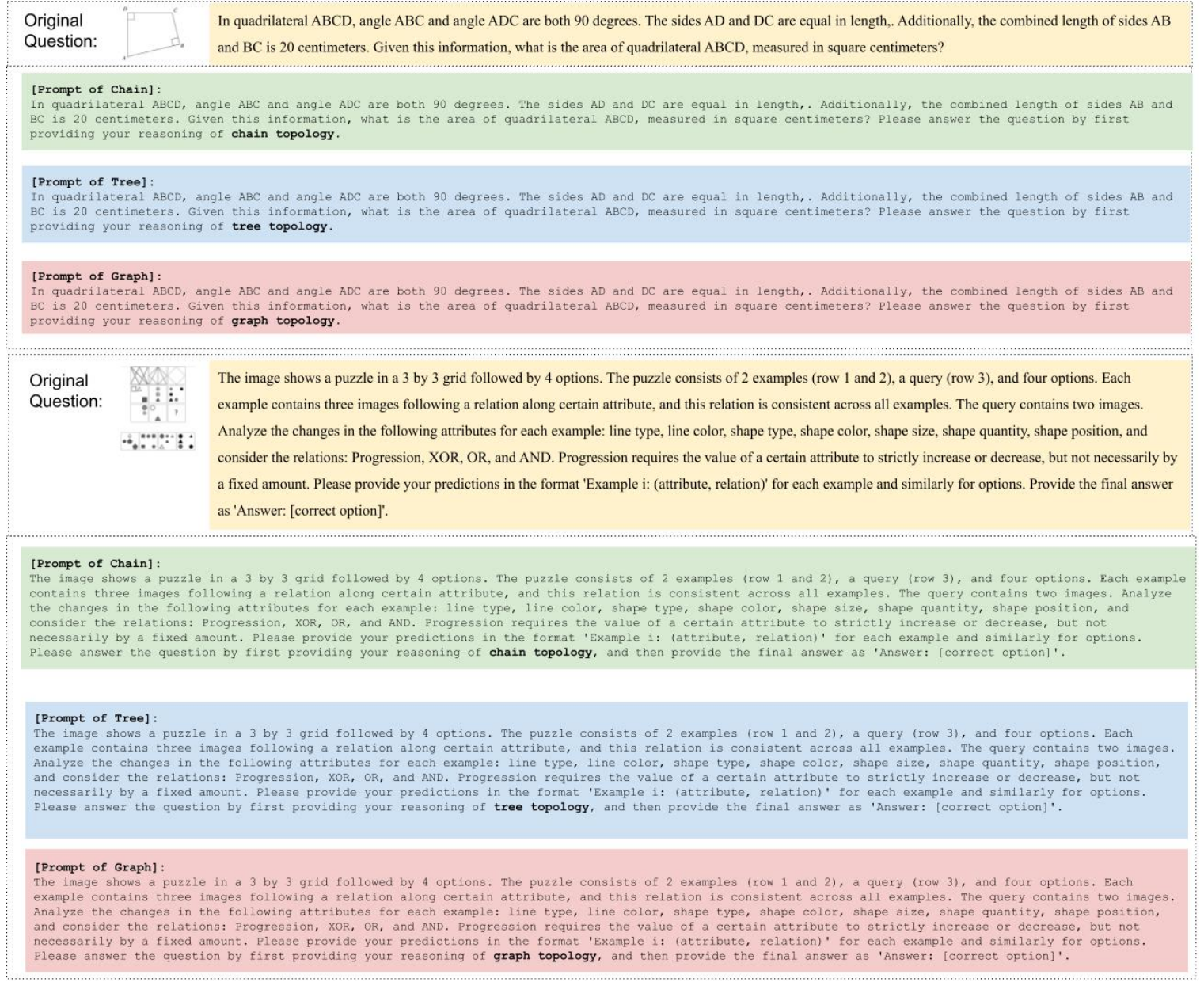}

    \vspace{0.5em} 

    \parbox{0.95\textwidth}{
        \centering
        \textit{Figure:} Prompts used in TopoAug to generate training samples for the MathVision (MATH-V) and VLM\_S2H datasets. For each question, we prompt the generation models to produce 10 to 20 responses across three distinct reasoning topologies—Chain, Tree, and Graph—with extensive flexibility in maximum depth, number of children and neighbors.
    }
\end{center}


\end{document}